\documentclass[letterpaper, 10 pt, conference]{ieeeconf}  % Comment this line out if you need a4paper

\IEEEoverridecommandlockouts                              % This command is only needed if 
\usepackage{fix-cm}
\usepackage{amsmath,amsfonts,bm}
\usepackage[table]{xcolor}
\usepackage{multirow}
\usepackage{array}
\usepackage{mathtools}
\usepackage{makecell}
\usepackage{multirow}
\usepackage{booktabs}
\usepackage{microtype}
\usepackage{graphicx}
\usepackage{float}
\newcolumntype{H}{>{\setbox0=\hbox\bgroup}c<{\egroup}@{}}

\newcommand{\E}{\mathbb{E}}
\newcommand{\R}{\mathbb{R}}

\newcommand{\boldA}{\boldsymbol{A}}
\newcommand{\boldB}{\boldsymbol{B}}

\newcommand{\boldF}{\boldsymbol{F}}
\newcommand{\boldG}{\boldsymbol{G}}
\newcommand{\boldH}{\boldsymbol{H}}
\newcommand{\boldI}{\boldsymbol{I}}
\newcommand{\boldL}{\boldsymbol{L}}

\newcommand{\boldW}{\boldsymbol{W}}
\newcommand{\boldX}{\boldsymbol{X}}

\newcommand{\bolda}{\boldsymbol{a}}
\newcommand{\bzero}{\boldsymbol{0}}

\newcommand{\boldh}{\boldsymbol{h}}
\newcommand{\boldl}{\boldsymbol{l}}
\newcommand{\bolds}{\boldsymbol{s}}

\newcommand{\boldx}{\boldsymbol{x}}
\newcommand{\boldy}{\boldsymbol{y}}
\newcommand{\boldz}{\boldsymbol{z}}

\newcommand{\calD}{\mathcal{D}}

\newcommand{\calL}{\mathcal{L}}

\newcommand{\calN}{\mathcal{N}}

\newcommand{\btheta}{\boldsymbol{\theta}}

\newcommand{\greencell}{\cellcolor{green!10}}

\title{\LARGE \bf
Uncertainty-Aware Adaptation of Large Language Models for Protein-Protein Interaction Analysis
}

\author{Sanket Jantre$^{1}$,  Tianle Wang$^{1}$, Gilchan Park$^{1}$, Kriti Chopra$^{1}$, Nicholas Jeon$^{2}$, \\ Xiaoning Qian$^{1,2}$, Nathan M. Urban$^{1}$, and Byung-Jun Yoon$^{1,2}$% <-this % stops a space
\thanks{\copyright This work has been submitted to the IEEE for possible publication. \protect\\ $^{1}$ Computing \& Data Sciences Directorate, Brookhaven National Laboratory. \protect\\ $^2$ Department of Electrical \& Computer Engineering, Texas A\&M University.}
}

\begin{document}
% \title{Uncertainty-Aware Adaptation of Large Language Models for Protein-Protein Interaction Analysis}

% \author{%
%   Sanket Jantre$^{1}$, Tianle Wang$^{1}$, Gilchan Park$^{1}$, Kriti Chopra$^{1}$, Nicholas Jeon$^{2}$,\\
%   Xiaoning Qian$^{1,2}$, Nathan M. Urban$^{1}$, and Byung-Jun Yoon$^{1,2}$%
%   \thanks{% 
%     \noindent $^{1}$ Computing \& Data Sciences Directorate, Brookhaven National Laboratory.\protect\\
%     $^{2}$ Department of Electrical \& Computer Engineering, Texas A\&M University.
%   }%
% }

\maketitle
\thispagestyle{empty}
\pagestyle{empty}

%%%%%%%%%%%%%%%%%%%%%%%%%%%%%%%%%%%%%%%%%%%%%%%%%%%%%%%%%%%%%%%%%%%%%%%%%%%%%%%%
\begin{abstract}

Identification of protein-protein interactions (PPIs) helps derive cellular mechanistic understanding, particularly in the context of complex conditions such as neurodegenerative disorders, metabolic syndromes, and cancer. Large Language Models (LLMs) have demonstrated remarkable potential in predicting protein structures and interactions via automated mining of vast biomedical literature; yet their inherent uncertainty remains a key challenge for deriving reproducible findings, critical for biomedical applications. In this study, we present an uncertainty-aware adaptation of LLMs for PPI analysis, leveraging fine-tuned LLaMA-3 and BioMedGPT models. To enhance prediction reliability, we integrate LoRA ensembles and Bayesian LoRA models for uncertainty quantification (UQ), ensuring confidence-calibrated insights into protein behavior. Our approach achieves competitive performance in PPI identification across diverse disease contexts while addressing model uncertainty, thereby enhancing trustworthiness and reproducibility in computational biology. These findings underscore the potential of uncertainty-aware LLM adaptation for advancing precision medicine and biomedical research.

% {\textbf{\textit{Clinical Relevance}}}\textemdash This is a brief statement on why a this might be of interest to practicing clinicians. Example: This establishes the anesthetic efficacy of 10\% intraosseous injections with epinephrine to positively influence cardiovascular function.

\end{abstract}


%%%%%%%%%%%%%%%%%%%%%%%%%%%%%%%%%%%%%%%%%%%%%%%%%%%%%%%%%%%%%%%%%%%%%%%%%%%%%%%%
\section{Introduction}
\label{sec:introduction}
Protein–protein interactions (PPIs) form the molecular foundation of cellular function, orchestrating everything from gene regulation and signal transduction to metabolic processes and immune response. The intricate network of these interactions, often termed the interactome, represents one of the most complex and dynamic systems in biology~\cite{Cusick2005}. Understanding this complex PPI network is particularly crucial in disease contexts, where aberrant protein interactions can lead to pathological states. Alterations in PPI networks have been implicated in numerous diseases, affecting fundamental cellular processes such as protein homeostasis, cell cycle regulation, and metabolic control. These disease-associated changes in the interactome can manifest through various mechanisms, from disrupted protein complex formation to altered signaling cascades, ultimately contributing to disease progression and severity. Elucidating these interaction networks is therefore essential for understanding disease mechanisms and developing therapeutic strategies \cite{ppis_disease}.

\begin{figure}
    \centering
    \includegraphics[width=0.825\linewidth]{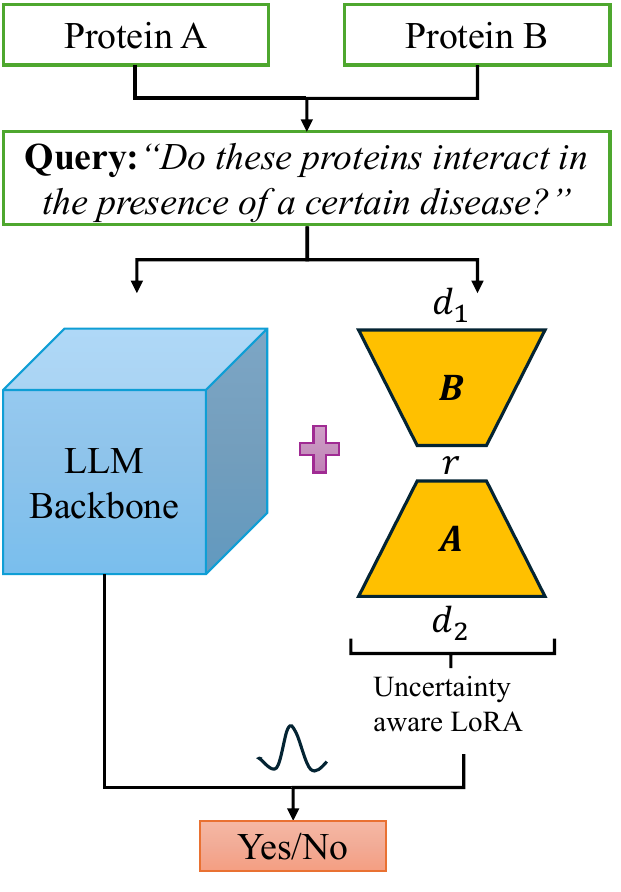}
    \caption{Illustration of our uncertainty-aware low-rank adaptation approach for pre-trained LLMs in protein-protein interaction prediction.}
    \label{fig:workflow}
\end{figure}

Traditional experimental methods for identifying PPIs, such as yeast two-hybrid screening and co-immunoprecipitation, have generated vast amounts of validated interaction data. These experimentally determined interactions have been systematically collected in comprehensive databases such as STRING~\cite{szklarczyk2023string}, BioGRID~\cite{oughtred2019biogrid}, and IntAct~\cite{orchard2014intact}, creating valuable resources for the research community. While experimental methods remain the gold standard for PPI validation, their labor- and time-intensive nature has led to the development of computational approaches to predict protein interactions. 

Early computational methods for PPI prediction primarily relied on sequence-based evolutionary patterns across species~\cite{weigt2009messagepassing, sanchezgarcia2019bipspi, chopra2020cornea}. As the field advanced, machine learning~(ML) approaches emerged, offering new ways to integrate multiple types of biological data~\cite{kewalramani2023state,zhang2024revolutionizing}. These included convolutional neural networks for analyzing protein sequence patterns, recurrent neural networks for capturing sequential dependencies, and graph neural networks for modeling the topology of protein interaction networks~\cite{soleymani2022ppi}.

The advancements of large language models~(LLMs)~\cite{shorinwa2024surveyuncertaintyquantificationlarge, minaee2024largelanguagemodelssurvey, radford2019language, brown2020languagemodelsfewshotlearners,openai2024gpt4technicalreport, grattafiori2024llama3herdmodels} have transformed multiple scientific domains through their unprecedented capabilities in understanding complex patterns and relationships in text data~\cite{Cui2024,luo2022biogpt,Singhal2023-zh,yang2022gatortronlargeclinicallanguage,bran2023chemcrowaugmentinglargelanguagemodels,taylor2022galacticalargelanguagemodel,wu2023bloomberggptlargelanguagemodel,thulke2024climategptaisynthesizinginterdisciplinary}. Building on their powerful language-processing capabilities, biology-specific models have been developed to tackle diverse tasks: ProtBERT~\cite{brandes2022proteinbert} and ESM3~\cite{hayes2025esm3} focus on protein sequence analysis, while BioGPT~\cite{luo2022biogpt} and BioMedGPT~\cite{luo2023biomedgpt} have shown promise in extracting biological knowledge from scientific literature. Recently, \cite{engel-park-2024-evaluating} has investigated PPI prediction using LLM-based approaches, demonstrating the potential of these models in specialized biomedical tasks. However, a critical challenge remains: such models often produce overly confident predictions, especially when trained on limited data, posing serious risks in high-stakes biomedical applications \cite{li2024thinktwicetrustingselfdetection, xiong2024llmsexpressuncertaintyempirical, leng2024tamingoverconfidencellmsreward, he2023preservingpretrainedfeatureshelps}. Uncertainty-aware adaptation of LLMs is particularly critical in biomedical applications where miscalibrated confidence levels could lead to erroneous conclusions about disease mechanisms or therapeutic targets. By incorporating principled uncertainty estimation and confidence calibration techniques, we can ensure that LLM-driven predictions are not only powerful but also reliable for guiding biomedical discoveries.

Within the broader machine learning community, uncertainty quantification has long been a challenge due to the high-dimensional parameter spaces of deep learning models. Traditional Bayesian methods relying on Markov chain Monte Carlo sampling often become intractable at scale \cite{Neal-1996, Izmailov-et-al-2021}. Consequently, approximate Bayesian methods such as variational inference~\cite{blundell2015weightuncertaintyneuralnetworks}, sparse learning \cite{jantre2023layer,jantre2023shrinkage}, and dimension reduction \cite{jantre2024active} have been explored. Alternately, ensemble-based methods like deep ensembles~\cite{lakshminarayanan2017simplescalablepredictiveuncertainty}, SWAG~\cite{maddox2019simple}, and SeBayS~\cite{jantre2022sequential} also provide principled strategies for capturing predictive variability. On the other hand, parameter-efficient fine-tuning techniques—particularly Low-Rank Adaptation~(LoRA)~\cite{hu2021loralowrankadaptationlarge}—makes it tractable to adapt large models for specific downstream tasks. To this end, novel UQ-aware fine-tuning approaches have emerged that combine LoRA with Bayesian or ensemble ideas, such as Bayesian LoRA \cite{yang2024bayesianlowrankadaptationlarge, wang2024blobbayesianlowrankadaptation, meo2024bloragates,onal2024gaussian} and LoRA ensembles \cite{wang2023loraensembleslargelanguage, balabanov2024uncertainty}.

In this study, we specifically focus on LLaMA-3 \cite{grattafiori2024llama3herdmodels} and BioMedGPT \cite{luo2023biomedgpt} as our primary LLM frameworks. We integrate LoRA-based fine-tuning with uncertainty-aware techniques to improve disease-specific PPI prediction. More concretely, we adopt Bayesian LoRA and LoRA ensemble methods to mitigate overconfidence and capture richer predictive variability. By leveraging the language-like structure of protein sequences, our approach naturally models intricate dependencies while generating well-calibrated estimates. To this end, our contributions include:
\begin{enumerate}
    \item Low rank adaptation-based fine-tuning of LLaMA-3-8B and BioMedGPT-LM-7B, comparing standard LoRA fine-tuning with Bayesian LoRA and LoRA ensembles to tackle disease-focused PPI prediction.
    \item Uncertainty quantification (UQ) integration to assess the reliability and robustness of PPI predictions.
    \item Comprehensive uncertainty-aware evaluation across disease-specific protein interaction networks, specifically those relevant to neurodegenerative disorders, metabolic diseases, and cancer.
\end{enumerate}
Overall, our results confirm that incorporating UQ strategies not only enhances PPI prediction accuracy but also yields better-calibrated confidence measures—critical for drawing robust conclusions in biomedical research. By advancing uncertainty-aware methods in LLM-based modeling, we lay the groundwork for safer, more reliable, and more informative computational tools in precision medicine.

\section{Preliminaries}
\label{sec:preliminaries}
Throughout the paper, all vectors and matrices are denoted by bold lowercase $(\boldl)$ and uppercase letters $(\boldL)$ respectively.

\subsection{Low-Rank Adaptation}
To adapt a pre-trained language model to downstream tasks, The authors of \cite{hu2021loralowrankadaptationlarge} introduced LoRA, a parameter-efficient fine-tuning approach. Assuming that weight changes exhibit a low intrinsic rank, LoRA optimizes rank decomposition matrices while keeping the pre-trained weights frozen.

Specifically, given that the weight update has a low-rank structure, the adapted forward pass is expressed as: 
\[
\boldh = (\boldW_0 + \Delta\boldW) \bolda = (\boldW_0 + \boldB\boldA) \bolda.
\] 
Here $\bolda$ and $\boldh$ represent the input and output vectors, respectively, of a large frozen pre-trained weight matrix $\boldW_0 \in \R^{d_1\times d_2}$. The matrices $\boldB\in\R^{d_1\times r}$ and $\boldA\in\R^{r\times d_2}$ contain trainable parameters, with $r \ll \min(d_1,d_2)$. This reduction in the number of parameters, allows LoRA to provide efficient fine-tuning with a decrease in storage requirements. We adopt this model as a baseline in our experiments.

\subsection{Bayesian model formulation}
Let $\calD=\{(\boldx_i,y_i)\}_{i=1,\cdots,N}$ be a training dataset with $N$ i.i.d. samples, where $\boldx$ represents the input samples and $y$ represents the output samples. Bayesian inference captures model uncertainty by inferring a probability distribution over model parameters, $\btheta = (\theta_1,\cdots,\theta_T) \in \R^T$, instead of learning a single deterministic model--$p(y|\boldx, \btheta)$. The posterior distribution follows Bayes’ rule: $p(\btheta|\calD) \propto p(\calD|\btheta) p(\btheta)$, where $p(\calD|\btheta)$ is the model likelihood and $p(\btheta)$ is the prior distribution. To make predictions for a new input $\boldx_{\rm new}$, Bayesian model averaging (BMA) is applied using the posterior distribution $p(\btheta|\calD)$ as follows:
\begin{align*}
p(y_{\rm new}|\boldx_{\rm new}, \calD) & = \int p(y_{\rm new}|\boldx_{\rm new}, \btheta) p(\btheta|\calD) d\btheta \\
& \approx \frac{1}{B} \sum_{b=1}^B p(y_{\rm new}|\boldx_{\rm new}, \btheta_b), \enskip \btheta_b \sim p(\btheta|\calD).
\end{align*}
This approach improves generalizability and model calibration by incorporating parameter uncertainty into predictions.

\section{Methodology}
\label{sec:methodology}
\subsection{LoRA Ensemble}
We employ an ensemble of LoRA models -- \emph{LoRA Ensemble} \cite{wang2023loraensembleslargelanguage, balabanov2024uncertainty} as an efficient strategy for uncertainty quantification in LLMs. Traditional deep ensembles yield better predictive performance and uncertainty estimation by training multiple models independently, but applying this directly to LLMs is infeasible due to high memory and computational costs.

To circumvent these issues, each LoRA Ensemble member fine-tunes the same pre-trained backbone $\boldW_0$ with a low-rank trainable modification $\Delta \boldW_m = \boldB_m \boldA_m$, where $\boldB_m \in \R^{d_1 \times r} $ and $\boldA_m \in \R^{r \times d_2}$ have significantly fewer parameters than the full model, $r_m \ll \min(d_1, d_2)$. These adapters are trained independently and in parallel, ensuring diverse solutions--$\{\boldW_1, \boldW_2, \dots, \boldW_M\}$. The ensemble prediction is computed by averaging outputs across $M$ ensemble members. For a given input $\boldx_{\rm new}$, if $y^m_{\rm new}$ represents the prediction from the $m$-th ensemble member, the final ensemble output (for continuous outcomes) is given by:
\[
p_{\rm ens} (y_{\rm new}|\boldx_{\rm new}) = \frac{1}{M} \sum_{m=1}^{M} p(y^m_{\rm new}|\boldx_{\rm new},\boldW_m).
\]
This approach retains the benefits of ensembling--improved accuracy, calibration, and robustness--while preserving efficiency by reusing the frozen backbone and only training lightweight LoRA adapters.

\subsection{Bayesian Low-Rank Adaptation}
Despite the availability of scalable posterior inference methods like variational inference \cite{blundell2015weightuncertaintyneuralnetworks}, a fully Bayesian treatment of LLMs remains computationally prohibitive. Instead, limiting Bayesian inference to LoRA parameters offers a more tractable means of capturing uncertainty in model predictions. However, even Markov chain Monte Carlo approaches can become excessively costly for inferring posteriors over the millions of LoRA parameters involved in large-scale models. As a practical compromise, \emph{Bayesian LoRA} \cite{yang2024bayesianlowrankadaptationlarge} employs the Laplace approximation to estimate the posterior over these low-rank parameters, centered around their \emph{maximum a posteriori} (MAP) estimate together with covariance equaling the Fisher information matrix \cite{daxberger2021laplace}.

To this end, let $\btheta$ denote the trainable LoRA parameters with a prior distribution of $\calN(\boldsymbol{0}, \lambda^{-1}\boldI)$. The Laplace approximation first calculates MAP estimate which is equivalent to maximizing the log-joint, $\log {\rm P} (\boldy, \boldX, \btheta)$
\begin{align*}
    \btheta_{\rm MAP} & = \underset{\btheta}{\arg\!\max} \log {\rm P}(\boldy,\boldX,\btheta) \\
    &= \underset{\btheta}{\arg\!\max} \log {\rm P}(\boldy|\boldX,\btheta) + \log {\rm P}(\btheta) \\
    &= \underset{\btheta}{\arg\!\max} \log {\rm P}(\boldy|\boldX,\btheta) + \frac{\lambda}{2} ||\btheta||^2_2 + {\rm const}
\end{align*}
where $\boldX$ represents the model inputs. The log prior term yields $L_2$-regularization on the trainable parameters, which can be incorporated into frequentist model training as a weight decay term with strength $\lambda/2$. Consequently, parameters from any model trained with an appropriate weight decay setting (e.g., via AdamW with its weight deca) can be directly reused.
% We can incorporate this in frequentist model training via weight decay term with $\lambda/2$ strength. As a result, parameters from any previously trained model that used a reasonable weight decay setting (for example, via AdamW with its weight decay) can be directly reused.

Next, to obtain an approximate posterior around $\btheta_{\rm MAP}$, Laplace method proceeds with a second-order Taylor expansion of the log-joint $\calL(\calD,\btheta)=\log p(\boldy, \boldX, \btheta)$ around $\btheta_{\rm MAP}$. Hence, by ignoring the higher-order terms, this yields
\[
    \calL (\calD,\btheta) \approx \calL (\calD,\btheta_{\rm MAP})
    + \frac{1}{2} (\btheta - \btheta_{\rm MAP})^\top \boldH \,(\btheta - \btheta_{\rm MAP}),
\]
where the first-order term zeros out due to the zero gradient at $\btheta_{\rm MAP}$ and $\boldH$ is the Hessian of the log-joint at $\btheta_{\rm MAP}$, $\nabla^2_{\btheta} \calL (\calD,\btheta)|_{\btheta_{\rm MAP}}$. Under this quadratic approximation,
\begin{equation}
\label{eqn:posterior}
    p(\btheta \mid \calD) \approx \calN \bigl(\btheta|\btheta_{\rm MAP},\, \boldH^{-1}\bigr).
\end{equation}
Hence, Laplace approximation turns out to be post-hoc Bayesian inference method which requires the additional step of computing the $H^{-1}$ matrix at $\btheta_{\rm MAP}$. In practice, computing the full Hessian $\boldH$ can be expensive, especially for large models due to quadratic complexity with respect to the number of model parameters. We use the positive semi-definite Fisher information matrix to circumvent the issue of the potentially indefinite Hessian, which arises when local convexity conditions fail to hold in large machine learning models. Accordingly, the Fisher information is defined by
\vspace{-0.2em}
\[
\boldF(\btheta) = \sum_{n}^N \E_{\hat{y}\sim{\rm P}(y|f_{\btheta}(\boldx_n))} \bigl[\boldG \boldG^\top \bigr]
\vspace{-0.2em}
\]
where $G=\nabla_{\btheta} {\rm P}(\hat{y}|f_{\btheta}(\boldx_n))$ represents the gradient and the expectation above is over the model’s output distribution. Next, in order to estimate the Fisher information in a manner that is both tractable and memory-efficient, we employ a \emph{Kronecker-Factored Approximate Curvature} (K-FAC) approach similar to \cite{ yang2024bayesianlowrankadaptationlarge}. In K-FAC, we treat Fisher as a block-diagonal matrix for each linear layer and factorize each block into two smaller matrices. For the $l$-th linear layer, we compute Fisher block $F_l$ using that layer's input activations $\bolda_{l-1}$ and log-likelihood gradients with respect to layer's pre-activation output $s_l$ denoted by $\boldG_{\bolds_l} = \nabla_{\bolds_l} \log {\rm P}(\boldy|\boldX,\btheta)$. Hence the expression is
\vspace{-0.2em}
\begin{equation}
\label{eqn:kfac}
   \boldF_l = \sum_{n=1}^N \E_{{\rm P}(y|f_{\btheta}(\boldx_n))} \bigl[ \bolda_{l-1}\bolda_{l-1}^\top \bigr] \otimes \E_{{\rm P}(y|f_{\btheta}(\boldx_n))} \bigl[ \boldG_{\bolds_l} \boldG_{\bolds_l}^\top \bigr]
\vspace{-0.2em}
\end{equation}
This approach avoids storing the full, dense Hessian, thereby reducing computational overhead.
By applying K-FAC to the LoRA parameters, we maintain a compact representation of uncertainty while keeping the overhead similar to standard training. However,
in Equation~(\ref{eqn:kfac}), the first expectation grows with the square of the layer’s input width, while the second grows with the square of the output width. Because LoRA adapters alternate between wide-input-narrow-output configuration and vice versa, one of these expectations can become especially large. To address this, we use an incremental singular value decomposition to factorize the large matrix into two new low-rank factors thereby saving memory. Further mathematical details are provided in Appendix~E of \cite{yang2024bayesianlowrankadaptationlarge}.

Once we infer the approximate posterior which is Gaussian as per Equation~\ref{eqn:posterior}, we can linearize the model predictions around the MAP estimate $\boldsymbol{\theta}_{\mathrm{MAP}}$ \cite{antoran2022adapting}. For a test input \(\mathbf{x}_{\rm new}\),
\vspace{-0.2em}
\[
f_{\btheta}(\boldx_{\rm new}) \approx f_{\btheta_{\rm MAP}}(\boldx_{\rm new}) + 
\nabla_{\btheta} f_{\btheta}(\boldx_{\rm new})\bigl|_{\btheta_{\rm MAP}}^\top
\bigl(\btheta - \btheta_{\rm MAP}\bigr).
\vspace{-0.2em}
\]
Because this expression is linear in \(\boldsymbol{\theta}\), integrating out 
the Gaussian posterior over \(\boldsymbol{\theta}\) yields a Gaussian predictive 
distribution for the logits:
\begin{align*}
f_{\btheta}(\boldx_{\rm new}) & \sim \calN \bigl(\boldy| f_{\btheta_{\rm MAP}}(\boldx_{\rm new}), \Lambda \bigr), \\
\text{wh}&\text{ere }  \Lambda  = \nabla_{\btheta} f_{\btheta_{\rm MAP}}(\boldx_{\rm new})^\top H^{-1} \nabla_{\btheta} f_{\btheta_{\rm MAP}}(\boldx_{\rm new}).
\end{align*}
Finally to efficiently sample from this predictive posterior, we use the Cholesky decomposition of 
$\Lambda = \mathbf{L} \mathbf{L}^\top$. Then,
\[
\hat{\boldy} = f_{\btheta}(\boldx_{\rm new}) = f_{\btheta_{\rm MAP}}(\boldx_{\rm new}) + \boldL \boldz, \quad
\boldz \sim \calN \bigl(\bzero, \boldI \bigr).
\]
This linearized predictive step, combined with a Gaussian approximate posterior, yields efficient uncertainty estimates in Bayesian LoRA approach for downstream tasks.

\begin{table*}[htb]
% \fontsize{7.05}{8.5}\selectfont
\fontsize{7.85}{11}\selectfont
\centering
\caption{PPI Prediction: The best results among all compared methods for a given LLM pre-trained model are highlighted in bold. All metrics are reported as means with standard deviations in subscript, based on three independent runs.}
\label{table:all-ppi}
\begin{tabular}{c|c|l|*{8}{l}} %*{1}{c}}
%========================================================================================
\toprule
LLM Model & Dataset & Methods & Acc ($\uparrow$) & NLL ($\downarrow$) & ECE ($\downarrow$) & Spec. ($\uparrow$) & Prec. ($\uparrow$) & F1 ($\uparrow$) & MCC ($\uparrow$) & AUROC ($\uparrow$) \\
\midrule
%=======================================================================================
\multirow{9}{*}{LLaMA-3} & \multirow{3}{*}{ND-PPI} & Single LoRA & 
87.42\textsubscript{\textcolor{gray}{1.64}} & 0.369\textsubscript{\textcolor{gray}{0.040}} & 0.089\textsubscript{\textcolor{gray}{0.023}} & 0.936\textsubscript{\textcolor{gray}{0.049}} & 0.885\textsubscript{\textcolor{gray}{0.007}} & 0.873\textsubscript{\textcolor{gray}{0.017}} & 0.758\textsubscript{\textcolor{gray}{0.025}} & 0.956\textsubscript{\textcolor{gray}{0.001}} \\
&& LoRA Ensemble & 
\textbf{89.84}\textsubscript{\textcolor{black}{0.30}} & \textbf{0.269}\textsubscript{\textcolor{black}{0.007}} & 0.054\textsubscript{\textcolor{gray}{0.002}} & \textbf{0.959}\textsubscript{\textcolor{black}{0.005}} & \textbf{0.905}\textsubscript{\textcolor{black}{0.002}} & \textbf{0.898}\textsubscript{\textcolor{black}{0.003}} & \textbf{0.802}\textsubscript{\textcolor{black}{0.005}} & \textbf{0.963}\textsubscript{\textcolor{black}{0.001}} \\
&& Bayesian LoRA &
86.87\textsubscript{\textcolor{gray}{0.73}} & 0.314\textsubscript{\textcolor{gray}{0.005}} & \textbf{0.031}\textsubscript{\textcolor{black}{0.005}} & 0.872\textsubscript{\textcolor{gray}{0.031}} & 0.869\textsubscript{\textcolor{gray}{0.008}} & 0.869\textsubscript{\textcolor{gray}{0.007}} & 0.738\textsubscript{\textcolor{gray}{0.015}} & 0.940\textsubscript{\textcolor{gray}{0.002}} \\
\cmidrule{2-11}
%=======================================================================================
& \multirow{3}{*}{M-PPI} & Single LoRA & 
85.82\textsubscript{\textcolor{gray}{0.26}} & 0.398\textsubscript{\textcolor{gray}{0.016}} & 
0.084\textsubscript{\textcolor{gray}{0.006}} & 0.908\textsubscript{\textcolor{gray}{0.036}} & 
0.863\textsubscript{\textcolor{gray}{0.007}} & 0.858\textsubscript{\textcolor{gray}{0.002}} & 
0.721\textsubscript{\textcolor{gray}{0.009}} & 0.937\textsubscript{\textcolor{gray}{0.002}} \\
&& LoRA Ensemble & 
\textbf{87.45}\textsubscript{\textcolor{black}{0.16}} & \textbf{0.308}\textsubscript{\textcolor{gray}{0.013}} & 
\textbf{0.051}\textsubscript{\textcolor{black}{0.010}} & 0.922\textsubscript{\textcolor{gray}{0.016}} & 
\textbf{0.878}\textsubscript{\textcolor{black}{0.003}} & \textbf{0.874}\textsubscript{\textcolor{black}{0.002}} & 
\textbf{0.752}\textsubscript{\textcolor{black}{0.004}} & \textbf{0.950}\textsubscript{\textcolor{black}{0.002}} \\  
&& Bayesian LoRA & 
83.41\textsubscript{\textcolor{gray}{1.17}} & 0.374\textsubscript{\textcolor{gray}{0.005}} & 
0.071\textsubscript{\textcolor{gray}{0.018}} & \textbf{0.932}\textsubscript{\textcolor{black}{0.038}} & 
0.850\textsubscript{\textcolor{gray}{0.003}} & 0.832\textsubscript{\textcolor{gray}{0.013}} & 
0.683\textsubscript{\textcolor{gray}{0.013}} & 0.925\textsubscript{\textcolor{gray}{0.004}} \\ 
\cmidrule{2-11}
%=======================================================================================
& \multirow{3}{*}{C-PPI} & Single LoRA & 
96.62\textsubscript{\textcolor{gray}{0.62}} & 0.094\textsubscript{\textcolor{gray}{0.011}} & 
0.033\textsubscript{\textcolor{gray}{0.002}} & 0.973\textsubscript{\textcolor{gray}{0.016}} & 
0.967\textsubscript{\textcolor{gray}{0.007}} & 0.966\textsubscript{\textcolor{gray}{0.006}} & 
0.932\textsubscript{\textcolor{gray}{0.013}} & 0.996\textsubscript{\textcolor{gray}{0.002}} \\ 
&& LoRA Ensemble & 
\textbf{97.86}\textsubscript{\textcolor{black}{0.00}} & \textbf{0.066}\textsubscript{\textcolor{black}{0.005}} & 
0.029\textsubscript{\textcolor{gray}{0.005}} & \textbf{0.980}\textsubscript{\textcolor{black}{0.000}} & 
\textbf{0.979}\textsubscript{\textcolor{black}{0.000}} & \textbf{0.979}\textsubscript{\textcolor{black}{0.000}} & 
\textbf{0.957}\textsubscript{\textcolor{black}{0.000}} & \textbf{0.997}\textsubscript{\textcolor{black}{0.000}} \\ 
&& Bayesian LoRA & 
96.97\textsubscript{\textcolor{gray}{1.24}} & 0.085\textsubscript{\textcolor{gray}{0.020}} & 
\textbf{0.027}\textsubscript{\textcolor{black}{0.002}} & 0.963\textsubscript{\textcolor{gray}{0.012}} & 
0.970\textsubscript{\textcolor{gray}{0.012}} & 0.970\textsubscript{\textcolor{gray}{0.012}} & 
0.940\textsubscript{\textcolor{gray}{0.025}} & 0.996\textsubscript{\textcolor{gray}{0.002}} \\ 
\midrule
%========================================================================================
%========================================================================================
\multirow{9}{*}{BioMedGPT} & \multirow{3}{*}{ND-PPI} & Single LoRA & 
86.90\textsubscript{\textcolor{gray}{0.99}} & 0.527\textsubscript{\textcolor{gray}{0.093}} & 0.101\textsubscript{\textcolor{gray}{0.013}} & 0.948\textsubscript{\textcolor{gray}{0.007}} & 0.879\textsubscript{\textcolor{gray}{0.008}} & 0.868\textsubscript{\textcolor{gray}{0.010}} & 0.747\textsubscript{\textcolor{gray}{0.018}} & 0.944\textsubscript{\textcolor{gray}{0.004}} \\
&& LoRA Ensemble & 
\textbf{88.94}\textsubscript{\textcolor{black}{0.75}} & 0.366\textsubscript{\textcolor{gray}{0.021}} & 0.073\textsubscript{\textcolor{gray}{0.011}} & \textbf{0.956}\textsubscript{\textcolor{black}{0.005}} & \textbf{0.897}\textsubscript{\textcolor{black}{0.006}} & \textbf{0.889}\textsubscript{\textcolor{black}{0.008}} & \textbf{0.785}\textsubscript{\textcolor{black}{0.013}} & \textbf{0.956}\textsubscript{\textcolor{black}{0.001}} \\
&& Bayesian LoRA &
87.17\textsubscript{\textcolor{gray}{0.36}} & \textbf{0.315}\textsubscript{\textcolor{black}{0.003}} & \textbf{0.031}\textsubscript{\textcolor{black}{0.003}} & 0.882\textsubscript{\textcolor{gray}{0.019}} & 0.872\textsubscript{\textcolor{gray}{0.004}} & 0.872\textsubscript{\textcolor{gray}{0.004}} & 0.744\textsubscript{\textcolor{gray}{0.008}} & 0.938\textsubscript{\textcolor{gray}{0.001}} \\
\cmidrule{2-11}
%=======================================================================================
& \multirow{3}{*}{M-PPI} & Single LoRA & 
85.99\textsubscript{\textcolor{gray}{0.44}} & 0.478\textsubscript{\textcolor{gray}{0.008}} & 0.092\textsubscript{\textcolor{gray}{0.003}} & 0.895\textsubscript{\textcolor{gray}{0.037}} & 0.863\textsubscript{\textcolor{gray}{0.007}} & 0.860\textsubscript{\textcolor{gray}{0.004}} & 0.722\textsubscript{\textcolor{gray}{0.011}} & 0.934\textsubscript{\textcolor{gray}{0.001}} \\
&& LoRA Ensemble & 
\textbf{87.45}\textsubscript{\textcolor{black}{1.20}} & \textbf{0.347}\textsubscript{\textcolor{black}{0.006}} & 0.053\textsubscript{\textcolor{gray}{0.001}} & \textbf{0.915}\textsubscript{\textcolor{black}{0.013}} & \textbf{0.877}\textsubscript{\textcolor{black}{0.011}} & \textbf{0.874}\textsubscript{\textcolor{black}{0.012}} & \textbf{0.751}\textsubscript{\textcolor{black}{0.023}} & \textbf{0.943}\textsubscript{\textcolor{black}{0.002}} \\
&& Bayesian LoRA &
83.37\textsubscript{\textcolor{gray}{0.74}} & 0.382\textsubscript{\textcolor{gray}{0.014}} & \textbf{0.033}\textsubscript{\textcolor{black}{0.011}} & 0.868\textsubscript{\textcolor{gray}{0.019}} & 0.836\textsubscript{\textcolor{gray}{0.006}} & 0.833\textsubscript{\textcolor{gray}{0.008}} & 0.669\textsubscript{\textcolor{gray}{0.013}} & 0.907\textsubscript{\textcolor{gray}{0.005}} \\
\cmidrule{2-11}
%=======================================================================================
& \multirow{3}{*}{C-PPI} & Single LoRA & 
97.68\textsubscript{\textcolor{gray}{0.82}} & 0.059\textsubscript{\textcolor{gray}{0.011}} & 
0.025\textsubscript{\textcolor{gray}{0.011}} & 0.976\textsubscript{\textcolor{gray}{0.006}} & 
0.977\textsubscript{\textcolor{gray}{0.008}} & 0.977\textsubscript{\textcolor{gray}{0.008}} & 
0.954\textsubscript{\textcolor{gray}{0.016}} & \textbf{0.998}\textsubscript{\textcolor{black}{0.001}} \\
&& LoRA Ensemble &  
\textbf{98.40}\textsubscript{\textcolor{black}{0.54}} & \textbf{0.052}\textsubscript{\textcolor{black}{0.000}} & 
\textbf{0.021}\textsubscript{\textcolor{black}{0.002}} & \textbf{0.980}\textsubscript{\textcolor{black}{0.000}} & 
\textbf{0.984}\textsubscript{\textcolor{black}{0.005}} & \textbf{0.984}\textsubscript{\textcolor{black}{0.005}} & 
\textbf{0.968}\textsubscript{\textcolor{black}{0.011}} & \textbf{0.998}\textsubscript{\textcolor{black}{0.000}} \\
&& Bayesian LoRA & 
\textbf{98.40}\textsubscript{\textcolor{black}{0.54}} & 0.064\textsubscript{\textcolor{gray}{0.005}} & 
0.031\textsubscript{\textcolor{gray}{0.001}} & 0.976\textsubscript{\textcolor{gray}{0.006}} & 
\textbf{0.984}\textsubscript{\textcolor{black}{0.005}} & \textbf{0.984}\textsubscript{\textcolor{black}{0.005}} & 
\textbf{0.968}\textsubscript{\textcolor{black}{0.011}} & \textbf{0.998}\textsubscript{\textcolor{black}{0.001}} \\
\bottomrule
%========================================================================================
\end{tabular}
\end{table*}

\section{Experimental Results}
\label{sec:expts_results}
In this section, we assess the performance of two uncertainty-aware LoRA adaptations---LoRA Ensemble and Bayesian LoRA---applied to LLaMA-3-8B and BioMedGPT-LM-7B models on publicly available protein-protein interaction datasets. As a baseline, we include a single LoRA model trained in a deterministic manner. All LoRA-based approaches were implemented using the PEFT library \cite{mangrulkar2022peft}, with each configuration run three times using different random seeds. We evaluate model performance and robustness by accuracy (Acc), negative log-likelihood (NLL), and expected calibration error (ECE) on the test sets. Additional details on the NLL and ECE metrics can be found in Appendix-A. Furthermore, we report Matthews Correlation Coefficient
(MCC), specificity (Spec.), precision (Prec.), F1-score, and Area under Receiver Operating Characteristic curve (AUROC) over test sets for a comprehensive view of predictive capabilities. Final metrics are summarized by the mean and standard deviation across three independent runs.

\vspace{1mm}
\noindent \textbf{PPI Datasets.}
The datasets analyzed in this work focus on PPIs relevant to various disease contexts, offering important insights into the underlying molecular mechanisms. 
We employ three primary balanced datasets—Neurodegenerative diseases PPI (ND-PPI), Metabolic disorders PPI (M-PPI), and Cancer PPI (C-PPI)-all of them containing equal numbers of positive and negative interactions.
% In particular, our study employs three primary datasets—Neurodegenerative diseases PPI (ND-PPI), Metabolic disorders PPI (M-PPI), and Cancer PPI (C-PPI)—all of which are balanced to contain an equal number of positive and negative interactions.

\textit{ND-PPI Dataset.} Adapted from \cite{pei2021predicting} and \cite{engel-park-2024-evaluating}, this dataset comprises 820 proteins involved in neurodegenerative diseases (e.g., Alzheimer’s, Parkinson’s, Huntington’s). Positive interactions are sourced from experimental and literature data, while negative pairs are generated by selecting protein pairs with no documented evidence of interaction based on established databases (e.g., DIP, BioGRID) with similar characteristics (e.g., similar functional annotations or expression profiles), resulting in 11,762 balanced interactions.

% \textit{ND-PPI Dataset.} The ND-PPI dataset, originally presented in \cite{pei2021predicting} and expanded in \cite{engel-park-2024-evaluating}, comprises 820 proteins implicated in neurodegenerative diseases such as Alzheimer’s, Parkinson’s, and Huntington’s. Positive interaction pairs were collected from experimentally validated interactions in public repositories and disease-specific literature. Negative pairs were then generated by selecting protein pairs with no documented evidence of interaction based on established databases (e.g., DIP, BioGRID), ensuring that these pairs shared comparable characteristics (e.g., similar functional annotations or expression profiles) to their positive counterparts. This procedure enabled the creation of a balanced dataset of 11,762 interactions (i.e., half positive and half negative), facilitating reliable model evaluation under neurodegenerative conditions.

\textit{M-PPI Dataset.} Also detailed in \cite{pei2021predicting} and \cite{engel-park-2024-evaluating}, the M-PPI dataset covers 1,063 proteins related to metabolic disorders (e.g., diabetes, obesity). As with the ND-PPI dataset, experimentally supported PPIs formed the positive set, while negative pairs were systematically identified based on the absence of known interactions. Consistent sampling methods ensure that protein abundance and disease relevance are preserved, resulting in a balanced set of 10,262 interactions.

% \textit{M-PPI Dataset.} Also reported in \cite{pei2021predicting} and further discussed in \cite{engel-park-2024-evaluating}, the M-PPI dataset focuses on metabolic disorders, featuring 1,063 proteins associated with conditions such as diabetes and obesity. As with the ND-PPI dataset, experimentally supported PPIs formed the positive set, while negative pairs were systematically identified based on the absence of known interactions. Sampling was performed in a manner that preserved overall protein abundance and disease relevance, resulting in 10,262 balanced interactions for metabolic disorder scenarios.

\textit{C-PPI Dataset.} Derived from \cite{qiu2021network}, the cancer PPI dataset originally had 933 positive and 1,308 negative interactions. Negative pairs were curated by excluding proteins with known interactions, and the set was balanced to 1,866 total interactions, in line with symmetric logistic matrix factorization approach \cite{pei2021predicting}.

% \textit{C-PPI Dataset.} The cancer PPI (C-PPI) dataset, derived from \cite{qiu2021network}, originally encompassed 933 positive and 1,308 negative interactions related to various oncological pathways. Negative pairs were curated by excluding any proteins with known interactions, thus ensuring high confidence in the non-interacting set. In alignment with prior methods \cite{pei2021predicting}, we balanced this dataset by randomly selecting negative pairs to match the number of positive pairs, culminating in a final, equal-sized set of 1,866 interactions. This approach follows established practice in symmetric logistic matrix factorization \cite{pei2021predicting}, which advocates balanced training samples for robust PPI prediction.

% Each dataset supports binary classification tasks (interaction present or absent) with an 80/20 train-test split, aligning with network-based PPI prediction frameworks \cite{qiu2021network} and recent large language model approaches \cite{engel-park-2024-evaluating}. This rigorous, balanced approach aims to improve computational PPI prediction across diverse disease contexts.

These datasets are employed to evaluate and advance computational models for predicting protein–protein interactions across distinct disease contexts by framing the tasks as binary (True/False) classification problems (see Fig.~\ref{fig:workflow}). Models assess whether proteins interact under specific conditions, contributing to our understanding of disease-specific interaction networks. Each dataset is partitioned into an 80\% training set and a 20\% testing set, consistent with network-based PPI prediction frameworks \cite{qiu2021network} and recent advances in large language model analysis under diverse biological conditions \cite{engel-park-2024-evaluating}. For further details on dataset selection, exploratory analyses, and additional context-specific investigations, readers are referred to \cite{engel-park-2024-evaluating} and related work.

% Collectively, these datasets enable binary (True/False) classification tasks, wherein models must ascertain the presence or absence of an interaction under disease-specific contexts (see Fig.~\ref{fig:workflow}). Each dataset is split into $80\%$ for training and $20\%$ for testing, and all methods are evaluated on these fixed test sets. This setup aligns with the framework of network-based PPI prediction \cite{qiu2021network} and leverages recent advances in large language model analysis under diverse biological conditions \cite{engel-park-2024-evaluating}. For further details on how individual protein types were selected and how additional exploratory analyses were performed—particularly regarding radiation exposure or disease-specific pathways—readers are referred to \cite{engel-park-2024-evaluating} and related work.

% By integrating these rigorously curated datasets and ensuring balanced representation of positive and negative interactions, our methodology aims to improve computational PPI prediction models across different disease contexts and elucidate the molecular interactions that drive these conditions.

\vspace{1mm}
\noindent \textbf{Implementation Details.}
In all experiments, we construct a LoRA ensemble using three individually fine-tuned LoRA learners. The LoRA matrices $\boldB$ are initialized to zero, while the entries of $\boldA$ follow a Kaiming Uniform initialization \cite{he-2015}. Optimization is performed using the AdamW optimizer with a learning rate of \(1\times 10^{-4}\), default hyperparameters, and a total of four training epochs. The batch size is set to $4$ for the ND-PPI and M-PPI cases and $16$ for the C-PPI case, following \cite{engel-park-2024-evaluating}. For Bayesian LoRA, the prior precision \(\lambda\) is fixed at $0.1$. 
We select the LoRA rank $(r)$ via hyperparameter search (Appendix-B) and report results for each LLM backbone and PPI dataset combination in Table~\ref{table:all-ppi} using the finalized $r$. Finally, LoRA adaptations are applied to the query, value, and output layer across all methods with \(\alpha=32\), a dropout rate of $0.05$, and a maximum sequence length of $50$.
% We perform hyperparameter search on the LoRA rank $(r)$ as detailed in the Apppendix~\ref{App:lora_rank} and report results in each LLM backbone and PPI dataset combination corresponding to the finalized choice of $r$ in Table~\ref{table:all-ppi}. Lastly, LoRA is applied to the queries, values, and output layer across all methods, with specific hyperparameters set \(\alpha=32\), a dropout rate of $0.05$, and a maximum sequence length of $50$.

\vspace{1mm}
\noindent \textbf{Results.} 
Table~\ref{table:all-ppi} summarizes the performance of the three LoRA-based models on ND-PPI, M-PPI, and C-PPI tasks using LLaMA-3 and BioMedGPT backbones.

In the ND-PPI prediction task, we demonstrate that the LoRA ensemble achieves the highest predictive accuracy among all models in both LLM settings and has the lowest NLL in the LLaMA-3 fine-tuning case. Conversely, Bayesian LoRA demonstrates the best calibration in both scenarios, exhibiting the lowest ECE and achieving the lowest NLL in the BioMedGPT fine-tuning case. Lastly, the LoRA ensemble reports the highest values for specificity, precision, F1-score, MCC, and AUROC among all models. In the M-PPI prediction task, we show that the LoRA ensemble achieves the highest predictive accuracy and lowest NLL in both LLM scenarios, while also attaining the lowest ECE in the LLaMA-3 case and highest specificity in BioMedGPT case. Conversely, Bayesian LoRA achieves the best calibration in the BioMedGPT case and the highest specificity in LLaMA-3 case. Finally, the LoRA ensemble outperforms all the models by achieving best precision, F1-score, MCC, and AUROC values.

In the C-PPI prediction task, we demonstrate that the LoRA ensemble once again achieves the highest predictive accuracy and lowest NLL in both settings, while also attaining the lowest ECE in the BioMedGPT scenario. Bayesian LoRA matches the best predictive accuracy in the BioMedGPT case and achieves the lowest ECE in the LLaMA-3 case. In the LLaMA-3 setting, the LoRA ensemble reports the highest values for specificity, precision, F1-score, MCC, and AUROC among all models. Additionally, it achieves the best specificity in the BioMedGPT case. Notably, both Bayesian LoRA and the LoRA ensemble attain the best precision, F1-score, and MCC values in the BioMedGPT case. Lastly, all three models yield identical AUROC values in the BioMedGPT case.

% We also conducted one‐sided Welch's t-tests to check whether the best-performing uncertainty-aware LoRA variant (Ensemble or Bayesian) significantly outperforms the baseline model of Single LoRA in accuracy (higher is better) or yields lower NLL and ECE (lower is better). Across most Llama-3 and BioMedGPT settings—especially on M-PPI and ND-PPI—these tests produced p-values below 0.05, indicating that the ensemble or Bayesian LoRA methods reliably improve predictive performance, uncertainty estimation, and calibration. A few comparisons including LLaMA-3 ND-PPI accuracy, BioMedGPT M-PPI accuracy, and BioMedGPT C-PPI all three metrics did not reach significance, suggesting that in those cases Single LoRA is statistically indistinguishable from the alternative. Overall, the results support that ensembling or Bayesian LoRA fine-tuning typically yields statistically significant gains over a single LoRA adapter.

We run one‐sided Welch’s t-tests to see if the best-performing uncertainty-aware LoRA variant (ensemble or Bayesian) outperforms Single LoRA in accuracy or lowers NLL/ECE. In most LLaMA-3 and BioMedGPT experiments—especially ND-PPI and M-PPI—the p-values are below 0.05, confirming significant gains; a handful of cases (LLaMA-3 ND-PPI accuracy, BioMedGPT M-PPI accuracy, and all BioMedGPT C-PPI metrics) are not significant, indicating comparable performance.

% These diagrams focus on the Llama-3 backbone—showing ND-PPI (top), M-PPI (middle), and C-PPI (bottom) calibration for Single LoRA, LoRA Ensemble, and Bayesian LoRA—while omitting BioMedGPT plots for brevity. They plot binned confidence versus observed accuracy and highlight the “gap” to reveal miscalibration. You’ll see that ensembling consistently yields the lowest ECE, with Bayesian LoRA usually close behind, both improving over the single-adapter baseline.

\begin{figure}
    \centering
    \includegraphics[width=0.995\linewidth]{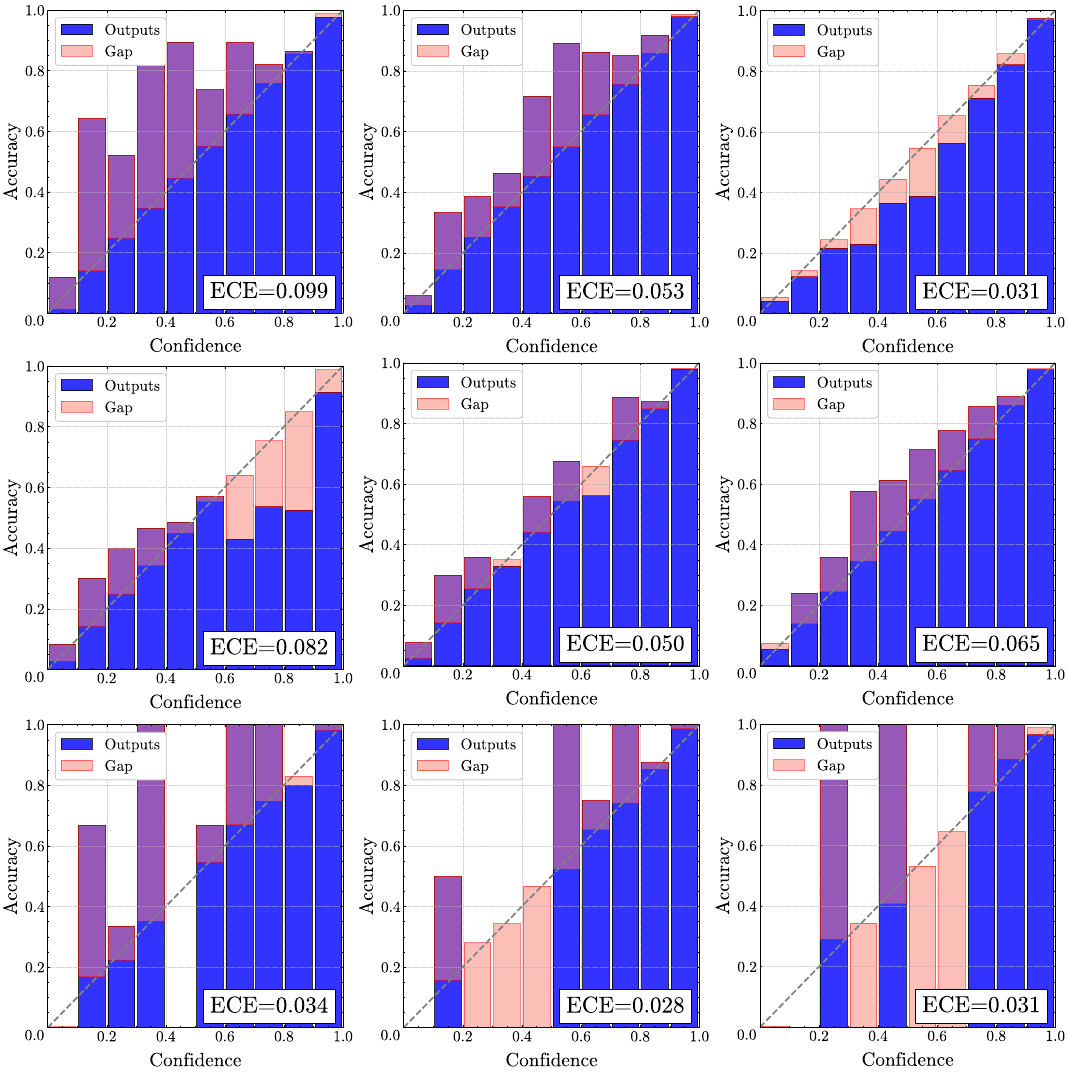}
    \caption{Reliability Diagrams: A visual representation of model calibration for the LLaMA-3 backbone trained on the ND-PPI (top row), M-PPI (middle row), and C-PPI (bottom row). Columns correspond, left to right, to the single LoRA, LoRA ensemble, and Bayesian LoRA models.}
    \label{fig:reliability_diagrams}
\end{figure}

Next, in Fig.~\ref{fig:reliability_diagrams} we show reliability diagrams for LLaMA-3 on ND-PPI (top row), M-PPI (middle row), and C-PPI (bottom row), comparing Single LoRA (left column), LoRA Ensemble (middle column), and Bayesian LoRA (right column). These diagrams plot binned confidence versus observed accuracy with the perfect calibration following the diagonal identity line, and any highlighted gap indicates miscalibration. Both LoRA ensemble and Bayesian LoRA variants consistently reduce this gap (and achieve lower ECE) relative to the single LoRA. We have omitted BioMedGPT plots for brevity.

\begin{table*}[htb]
% \fontsize{7.05}{8.5}\selectfont
\fontsize{7.25}{11}\selectfont
\centering
\caption{Effect of LoRA rank $(r)$ in single LoRA, LoRA ensemble, and Bayesian LORA models applied to Llama-3 and BioMedGPT across three PPI datasets. Final $r$ choices for each LLM backbone and dataset combination are shown by marking the corresponding results with light green color. Metrics are reported as means with standard deviations in subscript over three runs.}
\label{table:lora-rank-ablation-study}
\begin{tabular}{c|c|l|lll|lll|lll}
%=======================================================================================
\toprule
\multirow{2}{*}{LLM Model} & \multirow{2}{*}{Dataset} & \multirow{2}{*}{Methods} & \multicolumn{3}{c|}{Rank$=8$} & \multicolumn{3}{c|}{Rank$=16$} & \multicolumn{3}{c}{Rank$=32$} \\
\cmidrule{4-6}
\cmidrule{7-9}
\cmidrule{10-12}
&&&
Acc ($\uparrow$) & NLL ($\downarrow$) & ECE ($\downarrow$) & Acc ($\uparrow$) & NLL ($\downarrow$) & ECE ($\downarrow$) & Acc ($\uparrow$) & NLL ($\downarrow$) & ECE ($\downarrow$) \\
\midrule
%=======================================================================================
\multirow{9}{*}{LLaMA-3} & \multirow{3}{*}{ND-PPI} & Single LoRA & 
88.47\textsubscript{\textcolor{gray}{0.60}} & 0.330\textsubscript{\textcolor{gray}{0.036}} & 0.069\textsubscript{\textcolor{gray}{0.015}} & 
86.51\textsubscript{\textcolor{gray}{0.54}} & 0.362\textsubscript{\textcolor{gray}{0.036}} & 0.095\textsubscript{\textcolor{gray}{0.011}} &
\greencell 87.42\textsubscript{\textcolor{gray}{1.64}} & \greencell 0.369\textsubscript{\textcolor{gray}{0.040}} & \greencell 0.089\textsubscript{\textcolor{gray}{0.023}} \\
&& LoRA Ensemble & 
89.57\textsubscript{\textcolor{gray}{1.15}} & 0.278\textsubscript{\textcolor{gray}{0.022}} & 0.063\textsubscript{\textcolor{gray}{0.014}} & 
88.70\textsubscript{\textcolor{gray}{0.62}} & 0.302\textsubscript{\textcolor{gray}{0.002}} & 0.088\textsubscript{\textcolor{gray}{0.005}} &
\greencell 89.84\textsubscript{\textcolor{gray}{0.30}} & \greencell 0.269\textsubscript{\textcolor{gray}{0.007}} & \greencell 0.054\textsubscript{\textcolor{gray}{0.002}} \\
&& Bayesian LoRA &
85.14\textsubscript{\textcolor{gray}{1.45}} & 0.330\textsubscript{\textcolor{gray}{0.020}} & 0.060\textsubscript{\textcolor{gray}{0.017}} &
86.51\textsubscript{\textcolor{gray}{0.20}} & 0.317\textsubscript{\textcolor{gray}{0.003}} & 0.052\textsubscript{\textcolor{gray}{0.021}} &
\greencell 86.87\textsubscript{\textcolor{gray}{0.73}} & \greencell 0.314\textsubscript{\textcolor{gray}{0.005}} & \greencell 0.031\textsubscript{\textcolor{gray}{0.005}} \\
\cmidrule{2-12}
%=======================================================================================
& \multirow{3}{*}{M-PPI} & Single LoRA & 
85.84\textsubscript{\textcolor{gray}{0.35}} & 0.434\textsubscript{\textcolor{gray}{0.040}} & 0.089\textsubscript{\textcolor{gray}{0.012}} &
\greencell 85.82\textsubscript{\textcolor{gray}{0.26}} & \greencell 0.398\textsubscript{\textcolor{gray}{0.016}} & \greencell 0.084\textsubscript{\textcolor{gray}{0.006}} &
86.15\textsubscript{\textcolor{gray}{1.09}} & 0.515\textsubscript{\textcolor{gray}{0.072}} & 0.101\textsubscript{\textcolor{gray}{0.014}} \\
&& LoRA Ensemble & 
87.16\textsubscript{\textcolor{gray}{0.67}} & 0.334\textsubscript{\textcolor{gray}{0.012}} & 0.056\textsubscript{\textcolor{gray}{0.004}} &
\greencell 87.45\textsubscript{\textcolor{gray}{0.16}} & \greencell 0.308\textsubscript{\textcolor{gray}{0.013}} & \greencell 0.051\textsubscript{\textcolor{gray}{0.010}} &
87.50\textsubscript{\textcolor{gray}{0.75}} & 0.338\textsubscript{\textcolor{gray}{0.018}} & 0.055\textsubscript{\textcolor{gray}{0.005}} \\
&& Bayesian LoRA &
82.37\textsubscript{\textcolor{gray}{1.97}} & 0.392\textsubscript{\textcolor{gray}{0.038}} & 0.077\textsubscript{\textcolor{gray}{0.044}} &
\greencell 83.41\textsubscript{\textcolor{gray}{1.17}} & \greencell 0.374\textsubscript{\textcolor{gray}{0.005}} & \greencell 0.071\textsubscript{\textcolor{gray}{0.018}} &
81.78\textsubscript{\textcolor{gray}{2.06}} & 0.405\textsubscript{\textcolor{gray}{0.051}} & 0.093\textsubscript{\textcolor{gray}{0.046}} \\
\cmidrule{2-12}
%=======================================================================================
& \multirow{3}{*}{C-PPI} & Single LoRA & 
96.61\textsubscript{\textcolor{gray}{1.34}} & 0.116\textsubscript{\textcolor{gray}{0.024}} & 0.037\textsubscript{\textcolor{gray}{0.012}} &
\greencell 96.62\textsubscript{\textcolor{gray}{0.62}} & \greencell 0.094\textsubscript{\textcolor{gray}{0.011}} & \greencell 0.033\textsubscript{\textcolor{gray}{0.002}} &
96.61\textsubscript{\textcolor{gray}{0.31}} & 0.145\textsubscript{\textcolor{gray}{0.057}} & 0.034\textsubscript{\textcolor{gray}{0.006}} \\
&& LoRA Ensemble & 
97.33\textsubscript{\textcolor{gray}{0.54}} & 0.090\textsubscript{\textcolor{gray}{0.020}} & 0.036\textsubscript{\textcolor{gray}{0.005}} &
\greencell 97.86\textsubscript{\textcolor{gray}{0.00}} & \greencell 0.066\textsubscript{\textcolor{gray}{0.005}} & \greencell 0.029\textsubscript{\textcolor{gray}{0.005}} &
96.79\textsubscript{\textcolor{gray}{0.54}} & 0.081\textsubscript{\textcolor{gray}{0.019}} & 0.030\textsubscript{\textcolor{gray}{0.005}} \\
&& Bayesian LoRA &
96.61\textsubscript{\textcolor{gray}{0.82}} & 0.142\textsubscript{\textcolor{gray}{0.025}} & 0.034\textsubscript{\textcolor{gray}{0.008}} &
\greencell 96.97\textsubscript{\textcolor{gray}{1.24}} & \greencell 0.085\textsubscript{\textcolor{gray}{0.020}} & \greencell 0.027\textsubscript{\textcolor{gray}{0.002}} &
96.79\textsubscript{\textcolor{gray}{1.42}} & 0.119\textsubscript{\textcolor{gray}{0.040}} & 0.033\textsubscript{\textcolor{gray}{0.014}} \\
%=======================================================================================
\midrule
%=======================================================================================
\multirow{9}{*}{BioMedGPT} & \multirow{3}{*}{ND-PPI} & Single LoRA & 
86.29\textsubscript{\textcolor{gray}{1.58}} & 0.471\textsubscript{\textcolor{gray}{0.095}} & 0.095\textsubscript{\textcolor{gray}{0.022}} &
85.44\textsubscript{\textcolor{gray}{2.16}} & 0.539\textsubscript{\textcolor{gray}{0.053}} & 0.119\textsubscript{\textcolor{gray}{0.025}} &
\greencell 86.90\textsubscript{\textcolor{gray}{0.99}} & \greencell 0.527\textsubscript{\textcolor{gray}{0.093}} & \greencell 0.101\textsubscript{\textcolor{gray}{0.013}} \\
&& LoRA Ensemble & 
87.48\textsubscript{\textcolor{gray}{0.63}} & 0.403\textsubscript{\textcolor{gray}{0.065}} & 0.098\textsubscript{\textcolor{gray}{0.019}} &
88.00\textsubscript{\textcolor{gray}{1.19}} & 0.363\textsubscript{\textcolor{gray}{0.049}} & 0.087\textsubscript{\textcolor{gray}{0.022}} &
\greencell 88.94\textsubscript{\textcolor{gray}{0.75}} & \greencell 0.366\textsubscript{\textcolor{gray}{0.021}} & \greencell 0.073\textsubscript{\textcolor{gray}{0.011}} \\
&& Bayesian LoRA &
86.54\textsubscript{\textcolor{gray}{0.36}} & 0.326\textsubscript{\textcolor{gray}{0.007}} & 0.031\textsubscript{\textcolor{gray}{0.010}} &
86.82\textsubscript{\textcolor{gray}{0.60}} & 0.320\textsubscript{\textcolor{gray}{0.012}} & 0.033\textsubscript{\textcolor{gray}{0.007}} &
\greencell 87.17\textsubscript{\textcolor{gray}{0.36}} & \greencell 0.315\textsubscript{\textcolor{gray}{0.003}} & \greencell 0.031\textsubscript{\textcolor{gray}{0.003}} \\
\cmidrule{2-12}
%=======================================================================================
& \multirow{3}{*}{M-PPI} & Single LoRA & 
\greencell 85.99\textsubscript{\textcolor{gray}{0.44}} & \greencell 0.478\textsubscript{\textcolor{gray}{0.008}} & \greencell 0.092\textsubscript{\textcolor{gray}{0.003}} &
83.68\textsubscript{\textcolor{gray}{0.54}} & 0.542\textsubscript{\textcolor{gray}{0.026}} & 0.113\textsubscript{\textcolor{gray}{0.009}} &
84.95\textsubscript{\textcolor{gray}{0.43}} & 0.510\textsubscript{\textcolor{gray}{0.060}} & 0.101\textsubscript{\textcolor{gray}{0.008}} \\
&& LoRA Ensemble & 
\greencell 87.45\textsubscript{\textcolor{gray}{1.20}} & \greencell 0.347\textsubscript{\textcolor{gray}{0.006}} & \greencell 0.053\textsubscript{\textcolor{gray}{0.001}} &
87.14\textsubscript{\textcolor{gray}{1.39}} & 0.354\textsubscript{\textcolor{gray}{0.028}} & 0.062\textsubscript{\textcolor{gray}{0.010}} &
86.65\textsubscript{\textcolor{gray}{0.78}} & 0.357\textsubscript{\textcolor{gray}{0.003}} & 0.061\textsubscript{\textcolor{gray}{0.000}} \\
&& Bayesian LoRA &
\greencell 83.37\textsubscript{\textcolor{gray}{0.74}} & \greencell 0.382\textsubscript{\textcolor{gray}{0.014}} & \greencell 0.033\textsubscript{\textcolor{gray}{0.011}} &
83.29\textsubscript{\textcolor{gray}{0.57}} & 0.385\textsubscript{\textcolor{gray}{0.015}} & 0.037\textsubscript{\textcolor{gray}{0.018}} &
83.44\textsubscript{\textcolor{gray}{0.50}} & 0.374\textsubscript{\textcolor{gray}{0.005}} & 0.028\textsubscript{\textcolor{gray}{0.010}} \\
\cmidrule{2-12}
%=======================================================================================
& \multirow{3}{*}{C-PPI} & Single LoRA & 
97.86\textsubscript{\textcolor{gray}{0.54}} & 0.079\textsubscript{\textcolor{gray}{0.015}} & 0.026\textsubscript{\textcolor{gray}{0.002}} &
\greencell 97.68\textsubscript{\textcolor{gray}{0.82}} & \greencell 0.059\textsubscript{\textcolor{gray}{0.011}} & \greencell 0.025\textsubscript{\textcolor{gray}{0.011}} &
97.51\textsubscript{\textcolor{gray}{0.31}} & 0.063\textsubscript{\textcolor{gray}{0.005}} & 0.026\textsubscript{\textcolor{gray}{0.003}} \\
&& LoRA Ensemble & 
98.57\textsubscript{\textcolor{gray}{0.62}} & 0.063\textsubscript{\textcolor{gray}{0.009}} & 0.024\textsubscript{\textcolor{gray}{0.004}} &
\greencell 98.40\textsubscript{\textcolor{gray}{0.54}} & \greencell 0.052\textsubscript{\textcolor{gray}{0.000}} & \greencell 0.021\textsubscript{\textcolor{gray}{0.002}} &
98.40\textsubscript{\textcolor{gray}{0.00}} & 0.057\textsubscript{\textcolor{gray}{0.002}} & 0.024\textsubscript{\textcolor{gray}{0.001}} \\
&& Bayesian LoRA &
98.22\textsubscript{\textcolor{gray}{0.31}} & 0.072\textsubscript{\textcolor{gray}{0.009}} & 0.028\textsubscript{\textcolor{gray}{0.002}} &
\greencell 98.40\textsubscript{\textcolor{gray}{0.54}} & \greencell 0.064\textsubscript{\textcolor{gray}{0.005}} & \greencell 0.031\textsubscript{\textcolor{gray}{0.001}} &
97.86\textsubscript{\textcolor{gray}{0.54}} & 0.075\textsubscript{\textcolor{gray}{0.010}} & 0.031\textsubscript{\textcolor{gray}{0.007}} \\
\bottomrule
%=======================================================================================
\end{tabular}
\end{table*}

\section{Conclusion and Discussion}
In this study, we presented a novel uncertainty-aware adaptation of LLMs approach for predicting protein-protein interactions across multiple disease contexts. Leveraging fine-tuned LLaMA-3 and BioMedGPT models with LoRA ensemble and Bayesian LoRA, our approach consistently improved prediction accuracy, reliability, and robustness, as confirmed by comprehensive metrics such as negative log-likelihood and calibration error. LoRA ensembles excelled at achieving higher accuracy and reliable uncertainty estimates, while Bayesian LoRA provided well-calibrated predictions. Together, they demonstrated robustness in neurodegenerative, metabolic, and cancer-related PPI tasks. These findings underscore the benefits of incorporating principled uncertainty quantification into parameter-efficient fine-tuning for LLMs. To this end, our method can enable efficient UQ for large-scale AI models in diverse biological tasks, such as agentic AI workflows whose importance is growing in biological data analysis and drug discovery. Furthermore, the quantified uncertainty via our method could be used to assess the reliability of LLM outputs and optimize LLM prompts \cite{zhao2025pareto}, not only to improve the outputs but also their reliability.

Future work will examine more advanced LLM uncertainty-quantification techniques and extend this framework to broader biomedical applications. Potential directions include leveraging our approach to uncover disrupted protein interactions in disease pathways and, consequently, advance computational protein target discovery for therapeutic design.

\section*{APPENDIX}
\subsection{Robustness \& Predictive Uncertainty Evaluation Metrics}
% \label{app:implementation-baselines}
\label{app:uncertainty-metrics}
To assess model robustness and predictive uncertainty, we use Negative Log-Likelihood (NLL) and Expected Calibration Error (ECE). NLL evaluates how confidently a model predicts the correct labels. Given a test dataset \(\{\boldx_i, y_i\}_{i=1}^N\), NLL is:
\vspace{-0.3em}
\[
\text{NLL} = \frac{1}{N}\sum_{i=1}^{N} -\log {\rm P}_{\btheta}(y_n).
\vspace{-0.3em}
\]
A lower NLL indicates better confidence calibration, as overconfident incorrect predictions increase this value. On the other hand, ECE measures how well predicted confidence aligns with actual accuracy. Predictions are grouped into bins based on confidence, and ECE is calculated as:
\vspace{-0.5em}
\[
\text{ECE} = \sum_{m=1}^{M} \frac{|B_m|}{n} \left| \text{acc}(B_m) - \text{conf}(B_m) \right|.
\vspace{-0.5em}
\]
Here, \(\text{acc}(B_m)\) and \(\text{conf}(B_m)\) represent the average accuracy and confidence within bin \(B_m\), respectively:
\vspace{-0.4em}
\begin{equation*}
\begin{aligned}
    \text{acc}(B_m) &= \frac{1}{|B_m|} \sum_{i \in B_m} \mathbf{1}(\hat{y}_i = y_i), \\
    \quad \text{conf}(B_m) &= \frac{1}{|B_m|} \sum_{i \in B_m} P(\hat{y}_i),
\end{aligned}
\vspace{-0.4em}
\end{equation*}
where \(|B_m|\) is the number of samples in bin \(m\). Across all experiments, we set \(|B_m| = 15\).

\subsection{Effect of LoRA Rank}
\label{App:lora_rank}
In Table~\ref{table:lora-rank-ablation-study}, we present results of the hyperparameter search over the LoRA rank $r \in \{8, 16, 32\}$ for single LoRA, LoRA ensemble, and Bayesian LoRA across each LLM backbone and PPI dataset combination. For a fair comparison, we use the same selected rank $r$ across all three LoRA-based methods within each LLM backbone–PPI dataset combination. For both LLaMA-3 and BioMedGPT backbones, LoRA-based models achieve overall best performance on the ND-PPI dataset at $r=32$ and on the C-PPI dataset at $r=16$. In contrast, for the M-PPI dataset, LLaMA-3 and BioMedGPT backbones achieve overall best performance across LoRA-based models at $r=16$ and $r=8$, respectively.
% \vspace{-0.2em}

% \newpage
\section*{ACKNOWLEDGMENT}
\noindent This research was funded by the Biological and Environmental Research program in the US DOE's Office of Science under project B\&R\# KP1601017 and FWP\#CC140.
% \noindent This research was supported by the funding from the BER program in the US DOE's Office of Science under project B\&R\# KP1601017 and FWP\#CC140.
\vspace{-0.2em}

% Funded by the ASCR program in the US DOE’s Office of Science under grant 0000269227  and  projects B\&R\# KJ0402010 and FWP\# CC125

% The preferred spelling of the word "acknowledgment" in America is without an "e" after the "g". Avoid the stilted expression, "One of us (R. B. G.) thanks . . ."  Instead, try "R. B. G. thanks". Put sponsor acknowledgments in the unnumbered footnote on the first page.

% \addtolength{\textheight}{0.5cm}   % This command serves to balance the column lengths

%%%%%%%%%%%%%%%%%%%%%%%%%%%%%%%%%%%%%%%%%%%%%%%%%%%%%%%%%%%%%%%%%%%%%%%%%%%%%%%%

%%%%%%%%%%%%%%%%%%%%%%%%%%%%%%%%%%%%%%%%%%%%%%%%%%%%%%%%%%%%%%%%%%%%%%%%%%%%%%%%
%%%%%%%%%%%%%%%%%%%%%%%%%%%%%%%%%%%%%%%%%%%%%%%%%%%%%%%%%%%%%%%%%%%%%%%%%%%%%%%%

\end{document}